\documentclass[notitlepage,12pt]{jedm}
\usepackage{amsmath,amsfonts}
\usepackage{algorithmic}
\usepackage{array}
\usepackage[caption=false,font=normalsize,labelfont=sf,textfont=sf]{subfig}
\usepackage{textcomp}
\usepackage{stfloats}
\usepackage{url}
\usepackage{pifont}
\usepackage{verbatim}
\usepackage{graphicx}
\usepackage{multirow}

\usepackage{footnote}
\usepackage{enumitem}
\usepackage{longtable}
\usepackage{color}
\usepackage[normalem]{ulem}
\usepackage[table,dvipsnames]{xcolor}
\usepackage{adjustbox}
\usepackage{marvosym}
\usepackage{hyperref}
\usepackage{diagbox}
\usepackage[pagewise]{lineno}
\newcolumntype{M}[1]{>{\centering\arraybackslash}m{#1}}

\definecolor{effort}{HTML}{6DFCFF}
\definecolor{outcome}{HTML}{F4CCCC}
\hypersetup{
  colorlinks   = true, 
  urlcolor     = blue, 
  linkcolor    = blue, 
  citecolor   = blue 
}

\definecolor{royal_purple}{RGB}{153, 21, 78}

\makeatletter
\newcommand*{\addFileDependency}[1]{
  \typeout{(#1)}
  \@addtofilelist{#1}
  \IfFileExists{#1}{}{\typeout{No file #1.}}
}

\makeatother

\begin{document}

\title{Enhancing the De-identification of Personally Identifiable Information in Educational Data}
\date{} 


\newcommand{\authorFixedWidth}[1]{\parbox[t]{.5\textwidth}{\raggedright#1 \raisebox{0pt}[0pt][15pt]{}}}
\author{
\authorFixedWidth{{\large Zilyu Ji\thanks{These authors contributed equally to this work}\textsuperscript{\thefootnote}}\\Carnegie Mellon University\\Pittsburgh, USA\\zilyuj@andrew.cmu.edu}
\and
\authorFixedWidth{{\large Yuntian Shen\textsuperscript{\thefootnote}}\\Carnegie Mellon University\\Pittsburgh, USA\\yuntian2@andrew.cmu.edu}
\and
\authorFixedWidth{{\large Kenneth R. Koedinger}\\Carnegie Mellon University\\Pittsburgh, USA\\krk@cs.cmu.edu}
\and
\authorFixedWidth{{\large Jionghao Lin\thanks{Corresponding author: Jionghao Lin}}\\The University of Hong Kong\\Hong Kong, PR China\\jionghao@hku.hk}
}
\maketitle
\begin{abstract}
Protecting Personally Identifiable Information (PII), such as names, is a critical requirement in educational data mining to safeguard the privacy of students and teachers and maintain trust. Accurate PII detection is an essential step toward anonymizing sensitive information while preserving the utility of educational data. Motivated by recent advancements in artificial intelligence, our study investigates the GPT-4o-mini model as a cost-effective and efficient solution for PII detection tasks. We explore both prompting and fine-tuning approaches and compare GPT-4o-mini’s performance against established frameworks, including Microsoft Presidio and Azure AI Language. Our evaluation on two public datasets, CRAPII and TSCC, demonstrates that the fine-tuned GPT-4o-mini model achieves superior performance, with a recall of 0.9589 on CRAPII. Additionally, fine-tuned GPT-4o-mini significantly improves precision scores (a threefold increase) while reducing computational costs to nearly one-tenth of those associated with Azure AI Language. Furthermore, our bias analysis reveals that the fine-tuned GPT-4o-mini model consistently delivers accurate results across diverse cultural backgrounds and genders. The generalizability analysis using the TSCC dataset further highlights its robustness, achieving a recall of 0.9895 with minimal additional training data from TSCC. These results emphasize the potential of fine-tuned GPT-4o-mini as an accurate and cost-effective tool for PII detection in educational data. It offers robust privacy protection while preserving the data’s utility for research and pedagogical analysis. Our code is available on GitHub: \href{https://github.com/AnonJD/PrivacyAI}{https://github.com/AnonJD/PrivacyAI}\\ 

{\parindent0pt
\textbf{Keywords:} privacy, de-identification, personally identifiable information, large language models, fine-tuning GPT, cost-effectiveness, hidden in plain sight
}
\end{abstract}


\section{Introduction}\label{intro}

Personally Identifiable Information (PII) includes the information (e.g., names, email addresses, and phone numbers) that can identify an individual. Protecting PII is critical as computer-assisted learning systems become increasingly pervasive across all educational levels. For example, online human tutoring platforms collect vast amounts of interaction data, including conversational dialogues between students and tutors \cite{doi:10.1177/1477878518805308}. While the analysis of such educational data offers insights for data-informed research and improved pedagogical practices \cite{portnoff2021methods,chen2023building,2024.EDM-long-papers.20}, it also introduces significant privacy risks due to the sensitive nature of PII \cite{fung2010privacypreserving}. Safeguarding PII is essential to prevent unauthorized access and misuse \cite{10.1136/amiajnl-2012-001034}. Moreover, robust privacy protections are vital for fostering trust among educational stakeholders—students, educators, and parents—and ensuring the responsible adoption of learning technologies \cite{zeide2018education}. These efforts align with global regulatory frameworks, such as the General Data Protection Regulation (GDPR) and the Family Educational Rights and Privacy Act (FERPA), which mandate the protection of personal data.

Given the vast amounts of data collected by learning systems, automating the anonymization of PII in the education domain has become a critical necessity. Previous work has explored rule-based, statistical-based, and neural network-based approaches to detect PII in education-related tasks, such as essay grading \cite{chen2023building,liu2023geval}. Though showing effectiveness, given the advent of advanced AI models, PII detection can be further enhanced in detection accuracy, robustness, and generalizability when applied to diverse and complex educational datasets.
Moreover, many previous works \cite{caines-etal-2022-teacher,pal2024empirical,lison2021anonymisation,971193} focus solely on redacting PII from the original data (e.g., the name John from the original sentence \textit{``Thanks, John''}  processed by redacting: \textit{``Thanks, \texttt{[REDACTED]}''}). While this approach mitigates certain risks, it falls short of ensuring comprehensive privacy protection, especially given that current models still fail to achieve perfect accuracy in PII detection \cite{2024.EDM-posters.88}.  This raises concerns about potential PII leakage. Thus, these challenges underscore the urgent need for more advanced, scalable, and robust methods to facilitate the anonymization process in educational data, ensuring comprehensive protection of PII and mitigating the risks of unintended exposure.

As proposed by \citeN{10.1136/amiajnl-2012-001034}, the Hidden in Plain Sight (HIPS) method introduces the protection of PII in datasets by first identifying PII entities and then replacing them with synthetic information that retains contextual characteristics. Unlike traditional redaction methods that use generic tokens such as \texttt{[REDACTED]} to redact PII, the first step of the HIPS method replaces PII with tokens that indicate the category of information, such as \texttt{<Name>}, \texttt{<Email>}, or \texttt{<Address>}. This multi-category entity recognition is a prerequisite for HIPS, as a single, generic \texttt{[REDACTED]} tag—as used in other recent LLM-based redaction studies \cite{ctx11692351300004436}—lacks the semantic information required for generating realistic, type-specific surrogate data. The process of identifying PII entities relies on Named Entity Recognition (NER), a task in natural language processing that aims to locate and classify specific entities within text into predefined categories, such as names, email addresses, and phone numbers \cite{tjong2003introduction}. Once NER is performed on the data corpus, the original PII entities can be replaced with synthetic information while retaining their contextual relevance. For example, a sentence such as \textit{``\texttt{\{John Smith\}}\textsubscript{\texttt{Name}} lives at \texttt{\{123 Main Street\}}\textsubscript{\texttt{Address}}''} can be transformed into \textit{``\texttt{\{Alex Doe\}} lives at \texttt{\{456 Elm Avenue\}},''} preserving the sentence structure while anonymizing sensitive details. According to \cite{10.1093/jamia/ocz114}, the HIPS method enhances the protection of PII by minimizing the risk of re-identification through context-aware anonymization, while still maintaining the utility of the dataset for research and analysis. However, the effectiveness of the HIPS method also relies on the accuracy of the initial PII identification step performed through NER. Therefore, while the broader goal of our work is to improve the end-to-end de-identification of educational data, this study specifically focuses on the foundational detection stage, as its accuracy and categorical precision are necessary prerequisites for enabling high-utility de-identification methods like HIPS.

The recent advancements in large language models (LLMs) have introduced significant opportunities to enhance PII detection. Recent studies \cite{shreya_singhal_2024_12729884} have explored the use of LLMs, such as the GPT-4 model, to identify PII in education datasets. While these models exhibit high recall scores, indicating their ability to identify a broad range of PII entities, their performance often comes at a high computational and financial cost \cite{samsi2023wordswattsbenchmarkingenergy}. Moreover, these studies frequently report relatively low precision scores, which poses challenges for further data analysis. This trade-off is evident in the latest work; for example, \citeN{ctx11692351300004436} report that while prompted GPT-4o, LLaMA 3.3 70B, and LLaMA 3.1 8B all achieved high average recall scores (over 0.9), their precision remained an unsolved challenge, with even the top-performing GPT-4o only reaching 0.579. This persistent issue of low precision poses challenges for further data analysis. Although high recall is essential for safeguarding PII by ensuring that sensitive information is comprehensively identified, we argue that high precision is also important in the context of the HIPS method. Low precision, characterized by a high rate of false positives, can lead to unnecessary replacement of non-PII entities using the HIPS method. This can disrupt the semantic integrity of the dataset, potentially breaking important information relevant to further analysis of educational data. For example, consider the sentence, \textit{``The \texttt{\{Newton\}} method is used for optimization,''} which could be replaced with \textit{``The \texttt{\{David\}} method is used for optimization.''} In this case, the term \textit{``Newton''} was incorrectly identified as PII and replaced. Such disruptions might break the mathematical context and hinder further analysis.

To address these challenges, our study investigates methods to enhance the recall scores of PII detection while preserving precision and reducing computational costs. Striking this balance is crucial for developing a more cost-effective solution for large-scale anonymization without compromising robust privacy protection. To guide our investigation, we propose a main \textbf{Research Question (RQ)}: \textit{To what extent can large language model-based approaches effectively identify PII compared to baseline approaches, such as Microsoft Presidio and Azure AI Language?}

By answering this question, our study makes the following key contributions:

(1) \textbf{Fine-Tuned Model for PII Detection:} We employ and evaluate a fine-tuned GPT-4o-mini model for PII detection. Our findings show it achieves high recall scores (over 0.95) at a fraction of the cost of comparable commercial services.
       
(2) \textbf{Verifier Model for Enhancing Precision:} We introduce a two-step \textit{verifier model} that enhances detection precision. By reducing false positives, this approach helps better preserve non-PII data, which can support de-identification methods such as Hidden in Plain Sight (HIPS) that aim to minimize disruption to non-sensitive information.
    
(3) \textbf{Cultural and Gender Bias Analysis:} We conduct an analysis of model fairness, evaluating performance across cultural and gender groups using name-based subgroup analysis in the TSCC dataset, where names were replaced according to cultural and gender distributions. Our results reveal that the fine-tuned GPT-4o-mini delivers equitable performance, reducing the cultural biases present in established baseline models.
    
(4) \textbf{Evaluation on Domain Adaptation Study:} We demonstrate the model's robustness by testing it on a different dataset (i.e., TSCC). The model proves to be adaptable, achieving high accuracy (Precision of 0.9708, Recall of 0.9895, and $F_1$ Score of 0.9801) on the new domain after fine-tuning on a small sample, confirming its suitability for deployment across varied educational contexts.

\section{Related Work}
\subsection{Deidentification of Personally Identifiable Information}

A significant aspect of the deployment of learning technologies in the actual learning and teaching environment is ensuring data privacy. To address the concern about data privacy, two commonly used deidentification methods are direct redaction and Hidden In Plain Sight (HIPS) \cite{caines-etal-2022-teacher,osborne2022bratsynthetictextdeidentificationusing}. Both approaches aim to safeguard PII by concealing sensitive data but they differ significantly in their implementation and impact on downstream data usability. 

Redaction is one of the most widely used methods for PII protection in education-related research \cite{lison2021anonymisation,S_nchez_2015}. In this approach, sensitive information is replaced with generic placeholders, such as \textit{``[REDACTED]''}, effectively removing it from the dataset. For example, \textit{``John Smith contacted the office via john.smith@example.com''} can be redacted as \textit{``[REDACTED] contacted the office via [REDACTED].''} While redaction is effective at concealing sensitive information, maintaining consistently high accuracy in PII detection often comes at a significant cost. For instance, the study by \citeN{shreya_singhal_2024_12729884} employed prompting with the GPT-4 model, which incurs expenses of \$30.00 per 1M input tokens and \$60.00 per 1M output tokens. Given the vast amounts of data collected from learning systems, it is crucial to strike a balance between cost and performance to ensure practical scalability.

The HIPS method offers a more nuanced approach to protecting PII \cite{10.1136/amiajnl-2012-001034}. Instead of simply redacting sensitive information, HIPS replaces PII entities with synthetic but semantically equivalent placeholders that retain the category of the original information \cite{10.1093/jamia/ocz114}. Compared to the redaction method, HIPS could effectively enhance privacy protection. Even if some PII entities are missed during the initial detection step, their replacement with synthetic counterparts ensures no sensitive information is exposed. For example, \citeN{Carrell2020} employed the MITRE Identification Scrubber Toolkit (MIST) to identify PII entities, after which the HIPS method is used to replace those identified entities with realistic surrogates across two corpora. Following substitution, expert annotators examine the anonymized corpus to detect leaked PII—the entities that MIST failed to identify. On average, only 26.8\% of the leaked PII are detected, and 62.8\% of the entities considered leaked by human attackers are actually not leaked PII. In \citeN{osborne2022bratsynthetictextdeidentificationusing}, researchers introduced BRATsynthetic as a novel HIPS replacement strategy that leverages a Markov chain–based approach to dynamically substitute PII entities with realistic surrogates. This method notably reduces the risk of PII leakage due to false negatives in PII detection. For instance, under a 5\% false negative error rate, document-level leakage is decreased from 94.2\% (using a traditional HIPS replacement) to 57.7\%.

It should be noted that the effectiveness of the HIPS method relies heavily on the accuracy of the Named Entity Recognition (NER) step. Misidentified entities (false positives) could lead to unnecessary replacements, disrupting the dataset's integrity \cite{fung2010privacypreserving,carvalho2022surveyprivacypreservingtechniquesdata}. Conversely, undetected PII (false negatives) could leave sensitive information exposed. Thus, enhancing precision is important. However, existing work in the AI in education field that focuses on developing NER systems for downstream HIPS replacement still demonstrates low precision scores. For example, one study reported a precision score of 0.24 \cite{10.1007/978-981-19-5240-1_12}. Given the advantages of using the HIPS method compared to redaction, our study focuses on the application of the HIPS method and particularly aims to enhance the detection of PII entities.

 \subsection{Large Language Models for Data Privacy}
Recent advancements in LLMs, such as GPT-4, which are trained on extensive datasets from diverse domains, enable them to capture long-range dependencies and contextual nuances that are crucial for identifying PII. Studies leveraging GPT-based models have demonstrated their ability to achieve high recall scores, effectively identifying a broad range of PII entities \cite{wang2023gptnernamedentityrecognition}. Additionally, LLMs can leverage this internal knowledge to distinguish actual PII from non-PII. Some recent evidence supports this view: for instance, a recent study prompts GPT-3 for a NER task, suggesting that fine-tuning such large models could yield more notable results \cite{wang2023gptnernamedentityrecognition}, especially since fine-tuned LLMs tend to outperform their prompted counterparts on NLP tasks \cite{zhang2023machine,mosbach2023fewshot}. 

The approach of fine-tuning an LLM for PII identification aligns well with the emerging AI-for-education domain, where labeled text data for PII identification are often scarce and fragmented. Due to this limited availability, a model can be trained on data that are not representative of the actual use case, and less experienced users may require a tool that can quickly adapt to their specific domain. Consequently, a source model capable of learning effectively from sparse and unrepresentative datasets is needed. Previous work has shown that an LLM can be fine-tuned with just a few labeled examples (few-shot learning) \cite{mosbach2023fewshot}. Although studies also indicate that successful few-shot learning is achievable with other architectures, these typically require carefully designed learning strategies and meticulous parameter tuning to prevent overfitting \cite{shen2021partial,lee2023fewshot}. In contrast, larger LLMs have been proven to be more resistant to overfitting as their size increases \cite{mosbach2023fewshot,gadre2024language}, potentially owing to their memorization dynamics \cite{tirumala2022memorization}.

There has been work showing that current standard redactors perform worse in certain gender and cultural groups. The redactors identify names in the African and Asian / Pacific cultural group with a higher error rate \cite{mansfield2022maskdemographicbiasdetection}. Performance disparity has also been reported between identifying male and female names of an NER system \cite{zhao2024comprehensivestudygenderbias}. LLMs, which have been aligned to remedy gender and cultural bias, could be the base models for the development of a PII identification system with reduced bias \cite{ouyang2022traininglanguagemodelsfollow}.

A final advantage of fine-tuning an LLM for PII de-identification lies in its lower financial and computational cost compared to prompt-engineering an LLM for PII de-identification, which often requires multiple demonstrations in the input prompts \cite{wang2023gptnernamedentityrecognition}. The above observation highlights the potential for fine-tuning large models, with GPT emerging as a promising candidate. Several studies have utilized prompted GPT approaches to de-identify PII entities, demonstrating encouraging results \cite{shreya_singhal_2024_12729884,wang2023gptnernamedentityrecognition}. The most recent work by \citeN{ctx11692351300004436} provides a key benchmark, confirming high recall (over 0.9) for prompted models but also underscoring the limitations of low precision (0.579 for GPT-4o, 0.506 for LLaMA 3.3, and 0.262 for LLaMA 3.1) and calling for future research into fine-tuning and algorithmic bias analysis. However, the analysis of algorithmic bias has not been a primary focus in most prior PII de-identification studies, a gap our research aims to address. With the growing availability of cost-effective fine-tuning APIs and the lack of prior work using fine-tuned GPT for PII identification, we aim to explore this approach further.

\section{Methods}

\subsection{Data and Data Pre-processing}\label{sec_data}

Our study utilized the Cleaned Repository of Annotated Personally Identifiable Information (CRAPII)\footnote{Cleaned Repository of Annotated PII. \url{https://www.kaggle.com/datasets/langdonholmes/cleaned-repository-of-annotated-pii/data}} dataset \cite{2024.EDM-posters.88}, which comprises 22,688 samples of student writings collected from a massive open online course (MOOC) offered by a university in the United States.  The course focused on critical thinking through design, teaching learners strategies such as storytelling and visualization to solve real-world problems \cite{2024.EDM-posters.88}. The dataset includes seven PII categories as direct identifiers: \texttt{Names}, \texttt{Email Addresses}, \texttt{Usernames}, \texttt{IDs}, \texttt{Phone Numbers}, \texttt{Personal URLs}, and \texttt{Street Addresses} \cite{2024.EDM-posters.88}. A sample of the dataset is shown in Table \ref{tab:example_dataset}. In total, the dataset contains 4,871 labeled words categorized as PII entities. To enable entity-based matching during our analysis, we extracted the character-wise positions of all annotated PII entities within the text.

\begin{table}[htbp]
\centering
\caption{Illustrated Example of CRAPII Dataset}
\vspace{0.4cm}
\label{tab:example_dataset}
\renewcommand{\arraystretch}{1.3}
\resizebox{0.6\columnwidth}{!}{%
\begin{tabular}{|l|p{8cm}|}
\hline
\textbf{Attribute} & \textbf{Example Value} \\ \hline
\textbf{full\_text} & \textit{Hi John Doe. Tel: (555)555-5555} \\ \hline
\textbf{document} & 379 \\ \hline
\textbf{tokens} & [`Hi', `John', `Doe', `.', `Tel', `:', \\ &
`(555)555-5555'] \\ \hline
\textbf{labels} & 
\begin{tabular}[t]{@{}l@{}}[`O', `B-NAME', `I-NAME', `O', \\ `O', `O', `B-PHONE\_NUM']\end{tabular} \\ \hline
\textbf{trailing\_whitespace} & [True, True, False, True, False, \\ &
True, False] \\ \hline
\end{tabular}
}
\end{table}

We also introduced another dataset, the Teacher-Student Chatroom Corpus (TSCC) \cite{caines-etal-2022-teacher}, to examine the generalizability of our investigated models, as detailed in Section \ref{generalizability}. Generalizability is crucial for evaluating a model's robustness when applied to datasets with different distributions. The TSCC dataset contains 260 chatroom sessions, with a total of 41.4K conversational turns and 362K word tokens. We processed the dataset by extracting the \texttt{role} and \texttt{edited} columns and combining them into a simplified \texttt{role: text} format, where each conversational turn is represented on a new line. A sample excerpt of the processed transcription is shown below:

\begin{verbatim}
teacher: Hi there 〈STUDENT〉, all OK?  
student: Hi 〈TEACHER〉, how are you?  
\end{verbatim}

\subsection{Models for PII Detection}

\begin{figure*}[htbp]
\centerline{\includegraphics[scale=0.255]{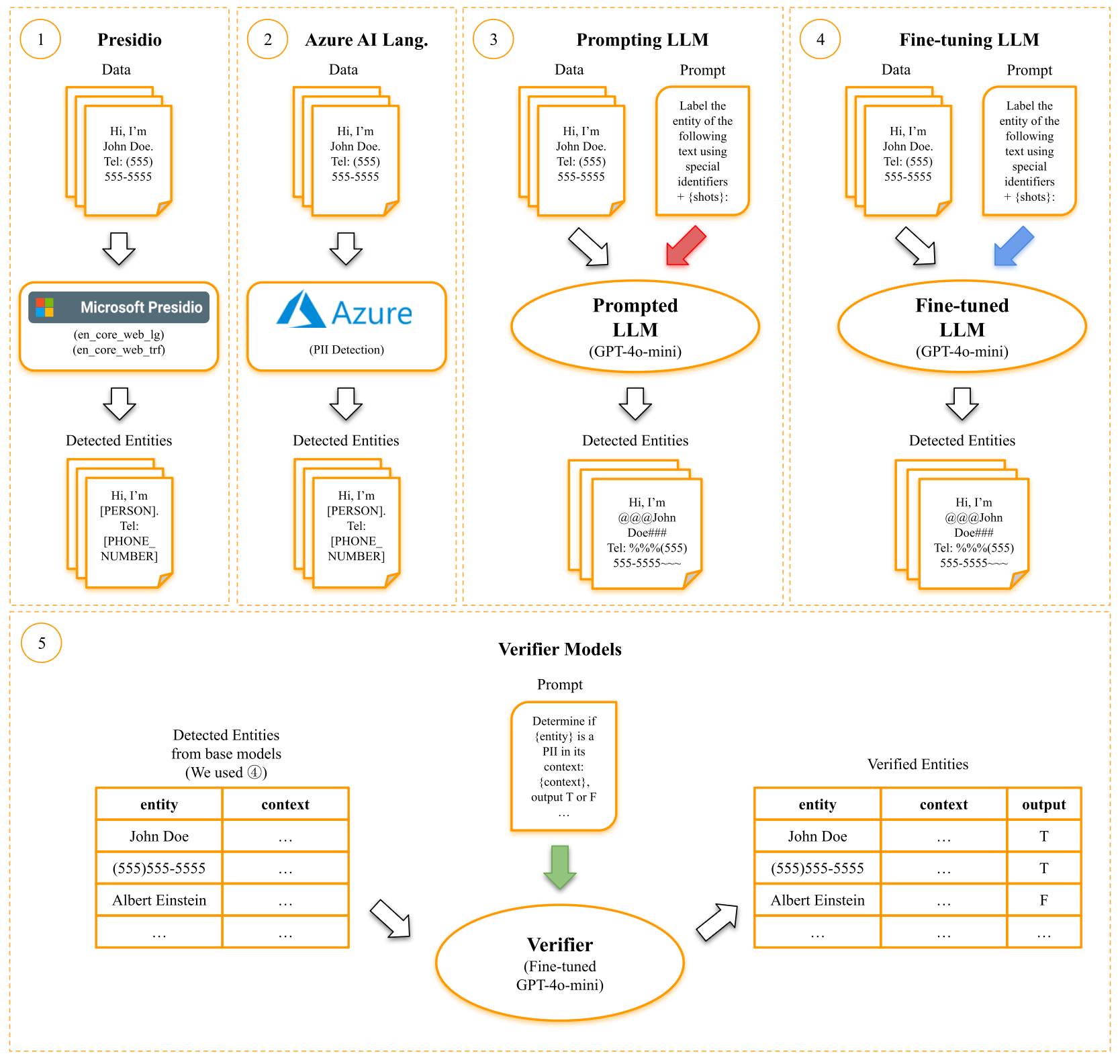}}
\caption{Overview of Five PII Detection Models.
\ding{172} \textbf{Presidio}: Uses Microsoft \textit{Presidio} with pre-trained \texttt{spaCy} models (\texttt{en\_core\_web\_lg} and \texttt{en\_core\_web\_trf}) to detect PII entities.
\ding{173} \textbf{Azure AI Language}: Leverages the PII detection feature in Microsoft \texttt{Azure AI Language} for entity recognition in input text.
\ding{174} \textbf{Prompting LLM}: Utilizes \texttt{GPT-4o-mini} with few-shot prompting and special identifiers to annotate PII entities in text. The red arrow shows the prompt used to guide the model for PII annotation.
\ding{175} \textbf{Fine-tuning LLM}: Fine-tunes a \texttt{GPT-4o-mini} model for PII detection. The blue arrow represents the prompt used to train the model during fine-tuning.
\ding{176} \textbf{Verifier Models}: Fine-tunes a \texttt{GPT-4o-mini} model to verify detected entities from the base model within their textual context. The green arrow indicates the prompt used to verify the entities within their context. Verification is performed with and without chain-of-thought (CoT) reasoning.
}
\Description{Diagram showing five PII detection models: Presidio, Azure AI Language, and Prompting LLM with arrows indicating workflows.}
\label{fig_diagram}
\end{figure*}

\textbf{Microsoft Presidio.}\footnote{Microsoft Presidio: Data Protection and De-identification SDK. \url{https://microsoft.github.io/presidio/}} It is an open-source toolkit designed for detecting and anonymizing PII across various text and image formats. 
It employs a regular-expression parser alongside NER models (such as \textit{en\_core\_web\_lg} and \textit{en\_core\_web\_trf}). Those NER models utilize linguistic patterns and deep learning techniques to classify words or phrases into predefined named entities (e.g., names, emails, and phone numbers). The backbone for \textit{en\_core\_web\_lg} is a convolutional neural network (CNN), and \textit{en\_core\_web\_trf} is a transformer-based model.\footnote{\url{https://spacy.io/models}} The model sizes for both are 382 and 436 MB. Additionally, \textit{Presidio} provides options for integrating external machine learning models for enhancing PII detection.

In our study, \textit{Presidio} serves as one of the baseline models for detecting PII entities. As indicated in previous work \cite{2024.EDM-posters.88}, \textit{en\_core\_web\_lg} is a standard NER model based on the \texttt{spaCy}\footnote{\url{https://spacy.io/}} framework that primarily relies on statistical techniques, while the transformer-based \textit{en\_core\_web\_trf} leverages pre-trained transformers, capturing long-range dependencies for more accurate context-based entity recognition. Thus, we primarily used these two configurations: \textit{en\_core\_web\_lg} and \textit{en\_core\_web\_trf}, as depicted in Fig. \ref{fig_diagram} (\ding{172}).

\textbf{Azure AI Language.}\footnote{Microsoft Azure AI Language: Personally Identifiable Information (PII) Detection. \url{https://learn.microsoft.com/en-us/azure/ai-services/language-service/personally-identifiable-information/overview}} It offers a cloud-based PII detection service capable of identifying and redacting sensitive information such as phone numbers and email addresses. Our study adopted \textit{Azure AI Language} as one of the baseline models for detecting PII entities in our dataset, as illustrated in Fig. \ref{fig_diagram} (\ding{173}). The version we use is \textit{2023-04-15-preview}. We used asynchronous processing through the REST API\footnote{Azure AI Language Service: Using Asynchronous Requests. \url{https://learn.microsoft.com/en-us/azure/ai-services/language-service/concepts/use-asynchronously}} to handle texts up to 125,000 characters per document. This approach is necessary because the longest transcript in our dataset contains 17,405 characters, exceeding the synchronous processing limit of 5,120 characters. Requests are batched with 5 documents per request and a rate limit of 1000 requests per minute, ensuring compliance with Azure’s service limits for PII detection.\footnote{Azure AI Language Service: Data Limits. \url{https://learn.microsoft.com/en-us/azure/ai-services/language-service/concepts/data-limits}} Each asynchronous request was completed within 24 hours, ensuring timely processing of all data.


\textbf{GPT-4o-mini (Prompting).} Motivated by the effectiveness of prompting LLMs in de-identifying PII, as demonstrated in a recent study \cite{shreya_singhal_2024_12729884}, our study adapts their prompting strategy with modifications to fit our task. Instead of employing \texttt{GPT-4}, as used in their work \cite{shreya_singhal_2024_12729884}, we opted for \texttt{GPT-4o-mini}, which requires only 1/200 of the cost per million input tokens and 1/100 of the cost per million output tokens compared to \texttt{GPT-4}.\footnote{OpenAI Models Pricing: \url{https://openai.com/api/pricing/}} Then, we modified the prompt structure by leveraging special identifiers for labeling detected entities rather than using a redaction method such as replacing detected PII with a generic label like \texttt{[REDACTED]} as shown in their work \cite{shreya_singhal_2024_12729884}.

Inspired by the GPT-NER method \cite{wang2023gptnernamedentityrecognition}, special identifiers can be used to mark entities, preserving non-PII content and ensuring precise PII detection. This method also reduces issues such as hallucinations and over-labeling that can arise with generative models. In particular, we require the detected entity positions to be an exact match to the true entity positions, but GPT often struggles with positional accuracy in long texts due to hallucinations and counting limitations \cite{ball2024countllmsfixedeffectfallacy}. To address this, we let GPT label detected PII entities by surrounding them with special identifiers for different categories (see Table~\ref{table:special_tokens}) and subsequently extract them using regular expressions, ensuring accurate detection and positioning.

\begin{table}[htbp]
\centering
\caption{Special Tokens for PII Detection}
\vspace{0.4cm}
\label{table:special_tokens}
\renewcommand{\arraystretch}{1.3}
\resizebox{0.6\columnwidth}{!}{%
\begin{tabular}{|l|c|l|}
\hline
\textbf{PII Category}  & \textbf{Special Identifiers}     & \textbf{Example} \\
\hline
\textbf{Student Name}           & \texttt{@@@}\textit{Text}\texttt{\#\#\#} & \texttt{@@@}\textit{John Doe}\texttt{\#\#\#} \\
\textbf{Personal URL}           & \texttt{\&\&\&}\textit{Text}\texttt{\$\$\$} & \texttt{\&\&\&}\textit{www.example.com}\texttt{\$\$\$} \\
\textbf{Personal Email}         & \texttt{QQQ}\textit{Text}\texttt{\textasciicircum\textasciicircum\textasciicircum} & \texttt{QQQ}\textit{johnd@example.com}\texttt{\textasciicircum\textasciicircum\textasciicircum} \\
\textbf{Phone Number}           & \texttt{\%\%\%}\textit{Text}\texttt{\textasciitilde\textasciitilde\textasciitilde} & \texttt{\%\%\%}\textit{(555)555-5555}\texttt{\textasciitilde\textasciitilde\textasciitilde} \\
\hline
\end{tabular}
}
\end{table}

To guide \texttt{GPT-4o-mini} in accurately identifying different types of PII entities, we employ a few-shot learning strategy that incorporates robust text-preservation instructions adapted from \citeN{shreya_singhal_2024_12729884} and tailored to our multi-class labeling task. Few-shot learning provides contextual examples that help the model understand the diversity of entity types and nuances within our dataset \cite{brown2020languagemodelsfewshotlearners}. This approach is chosen because the additional examples often improve the model's ability to handle ambiguous cases and enhance consistency in PII identification. Table \ref{table:prompt_structures} outlines the structure of the \textbf{System}, \textbf{User}, and \textbf{Assistant} prompts used in prompting \texttt{GPT-4o-mini}. Three examples\footnote{\url{https://github.com/AnonJD/PrivacyAI/blob/main/few-shots-example.txt}} are selected from the CRAPII dataset and incorporated into the user prompt. The text-preservation instructions adapted from \citeN{shreya_singhal_2024_12729884} are highlighted in bold for clarity. The \textbf{Assistant}'s output is the input text with labeled PII entities. The prompting process is detailed in Fig. \ref{fig_diagram} (\ding{174}).

\begin{table}[htbp]
\centering
\caption{Prompt Structures for Prompting-GPT-4o-mini Using Special Identifiers}
\vspace{0.4cm}
\label{table:prompt_structures}
\renewcommand{\arraystretch}{1.3}
\resizebox{0.85\columnwidth}{!}{%
\begin{tabular}{|c|p{15cm}|}
\hline
\textbf{Role} & \textbf{Content} \\
\hline
\textbf{System} & \textit{You are an expert in labeling Personally Identifiable Information (PII). Start your response right away without adding any prefix (such as ``Response:'') or suffix.} \\
\hline
\textbf{User} & \textit{Label the entity of the following text:} \texttt{@@@}, \texttt{\#\#\#} \textit{to label student name;} \texttt{\textnormal{\&\&\&}}, \texttt{\$\$\$} \textit{to label personal URL;} \texttt{QQQ}, \texttt{\textasciicircum\textasciicircum\textasciicircum} \textit{to label personal email;} \texttt{\%\%\%}, \texttt{\textasciitilde\textasciitilde\textasciitilde} \textit{to label phone number. \textbf{Ensure that the rest of the text remains unchanged, word for word. Maintain the original punctuation, quotation marks, spaces, and line breaks. If the text does not contain any PII, return it as is.}} \\
& \textbf{\textit{For example, if the input is:}} \texttt{\{Example One\}} \\ &
\textbf{\textit{The output should be:}} \texttt{\{Example One with Labeled PII\}}
\textit{Another example:} \texttt{\{Example Two\}}
\textit{Another example:} \texttt{\{Example Three\}}
\textit{Please repeat this process with the following file:} \texttt{\{Text Input\}} \\
\hline

\textbf{Assistant} & 
\texttt{\{Text Input with Labeled PII\}}
\\
\hline
\end{tabular}
}
\end{table}

\textbf{GPT-4o-mini (Fine-tuning).} Our study employed a fine-tuning strategy on \texttt{GPT-4o-mini} for PII detection. We utilized the same approach of incorporating special tokens, as shown in Table~\ref{table:special_tokens}. Then, \texttt{GPT-4o-mini} was fine-tuned on the \textbf{Base Train Set} and evaluated on the \textbf{Test Set} (as defined in Section \ref{split_data}) to assess its performance. 
The detailed structure of the system, user, and assistant prompts used during the fine-tuning process is presented in Table~\ref{table:fine_tuning_prompt_structures}, while the implementation process is illustrated in Fig.~\ref{fig_diagram} (\ding{175}).

\begin{table}[htbp]
\centering
\caption{Details of Fine-tuning GPT-4o-mini Using Special Identifiers}
\vspace{0.4cm}
\label{table:fine_tuning_prompt_structures}
\renewcommand{\arraystretch}{1.3}
\resizebox{0.6\columnwidth}{!}{%
\begin{tabular}{|c|p{10cm}|}
\hline
\textbf{Role} & \textbf{Content} \\
\hline
\textbf{System} & \textit{You are an expert in labeling Personally Identifiable Information. Start your response right away without adding any prefix (such as Response:) or suffix.} \\
\hline
\textbf{User} & \textit{Label the entity of the following text: @@@, \#\#\# to label student name; \textnormal{\&\&\&}, \$\$\$ to label personal URL; QQQ, \textasciicircum\textasciicircum\textasciicircum to label personal email; \%\%\%, \textasciitilde\textasciitilde\textasciitilde to label phone number.} \\ &
\texttt{\{Text Input\}}\\
\hline
\textbf{Assistant} & \texttt{\{Text Input with Labeled PII\}} \\
\hline
\end{tabular}
}
\end{table}

\textbf{Verifier Models.} To improve the precision (Equation \ref{eq:precision}) of PII detection while maintaining recall (Equation \ref{eq:recall}), we propose the addition of a verifier model as a second step to verify whether predictions are accurately identified as PII. This approach is inspired by recent studies \cite{brandfonbrener2024vermctssynthesizingmultistepprograms,li2023makinglargelanguagemodels} that integrate verifiers into multi-step reasoning tasks in LLMs. For the implementation of verifier models, we propose two variants: \textit{Verifier Model I (Without CoT)} and \textit{Verifier Model II (With CoT)}. These verifier models assess detected entities within their surrounding context to eliminate false positives while retaining true PII entities, thereby enhancing the precision of PII detection systems. Both verifiers are fine-tuned versions of the \texttt{GPT-4o-mini} model. The process of the verifier model approach is illustrated in Fig.~\ref{fig_diagram} (\ding{176}).

The \textit{Verifier Model II (With CoT)} incorporates Chain-of-Thought (CoT) reasoning, a method shown to enhance decision-making by breaking down complex problems into intermediate steps \cite{wei2023chainofthoughtpromptingelicitsreasoning}. The \textit{Verifier Model II (With CoT)} aims to improve interpretability and robustness by generating reasoning before the final classification. However, it generates more output tokens, leading to higher computational costs. The \textit{Verifier Model I (Without CoT)}, which directly classifies entities without reasoning, is retained for scenarios where computational resources are limited or speed is a higher priority.


To construct the training data for the verifier models, we used the dedicated \textbf{Verifier Train Set} (as defined in Section 3.3). The process begins by running a base PII detection model of the user's choice on this dataset to generate an initial list of detected PII entities. For each detected entity from this list, we then extracted it along with its surrounding textual context. This \texttt{(entity, context)} pair was subsequently labeled as \textbf{T} (True PII) or \textbf{F} (False PII) by comparing the entity to the ground-truth labels.

For the \textit{Verifier Model II (With CoT)}, we use \texttt{GPT-4o-mini} to generate reasoning to support the classification. If the reasoning does not align with the ground truth after six attempts, the label defaults to \texttt{T} to avoid mistakenly removing true PII entities and prioritize privacy preservation. The prompt structures for both verifier models are shown in Table~\ref{table:verifier_prompts}.

\begin{table*}[htbp]
\centering
\caption{Prompt Structures for Verifier Models}
\vspace{0.4cm}
\label{table:verifier_prompts}
\renewcommand{\arraystretch}{1.5}
\resizebox{0.75\textwidth}{!}{
\begin{tabular}{|c|p{8.5cm}|p{8.5cm}|}
\hline
\textbf{\large Role} & \textbf{\large Verifier Model I (Without CoT)} & \textbf{\large Verifier Model II (With CoT)} \\
\hline
\textbf{\large System} & \multicolumn{2}{p{16cm}|}{\large \textit{You are an expert in labeling Personally Identifiable Information. Start your response right away without adding any prefix (such as Response:) or suffix.}} \\ \hline
\textbf{\large User} & \large \textit{Determine if \texttt{\{entity\}} is a privately identifiable information in its context: \texttt{\{context\}}, think carefully before saying no to protect against PII leakage, only output T or F.} & \large \textit{Determine if \texttt{\{entity\}} is a privately identifiable information in its context: \texttt{\{context\}}. Think step-by-step before outputting T or F, format your response as (your reasoning) + [Response:] T or F} \\
\hline
\textbf{\large Assistant} & \large \texttt{``T'' or ``F''} & \large \texttt{``\texttt{\{CoT Reasoning\} + T}'' or ``\texttt{\{CoT Reasoning\} + F}''} \\
\hline
\end{tabular}
}
\end{table*}

Once the verifier models are trained, they can be applied to verify detected entities from any base model depending on user priorities. It is important to note that applying the verifier will not increase recall, as false negatives remain unchanged. Instead, the verifier reduces false positives, potentially at the expense of some true positives. If preserving PII is a higher priority than maximizing precision, the verifier should be applied to the base model with the highest recall. Conversely, for tasks that emphasize precision, the verifier may be applied to other base models as needed.

\textbf{GPT-4o-mini Training and Inference.} We set \textit{temperature} = 0 and \textit{top\_p} = 1 for both our GPT-4o-mini-based PII detector models (prompted and fine-tuned) and all verifier models to ensure strict adherence to the required output formats as suggested in the prior research \cite{wang2023gptnernamedentityrecognition}. For the fine-tuning process, we adopted the hyperparameters recommended by the OpenAI platform: \textit{epochs} = 2, \textit{batch\_size} = 1, and a \textit{learning\_rate\_multiplier} = 1.8. No further hyperparameter tuning was performed due to these platform constraints.

\subsection{Splitting Data for Training and Testing}\label{split_data}
Our study focuses on four specific categories from the CRAPII dataset for PII detection: \texttt{NAME}, \texttt{URL\_PERSONAL}, \texttt{EMAIL}, and \texttt{PHONE\_NUM}. 
To support our multi-stage experimental design, which involves training base models and then separate verifier models, the 22,688 files in the CRAPII dataset are split into three distinct sets: \textbf{Base Train Set} (25\%, 5,672 files), \textbf{Verifier Train Set} (15\%, 3,403 files), and \textbf{Test Set} (60\%, 13,613 files). This split ensures that all sets contain a sufficient number of entities from each category, including rare categories such as \texttt{PHONE\_NUM}, which only has 15 entities in total. Table~\ref{tab-entity_counts} presents the distribution of true entity counts across the three sets.

\begin{table}[htbp]
\centering
\caption{True Entity Counts Across Data Splits}
\vspace{0.4cm}
\label{tab-entity_counts}
\renewcommand{\arraystretch}{1.3}
\resizebox{0.6\textwidth}{!}{%
\begin{tabular}{|l|r|r|r|r|}
\hline
\textbf{Entity Type} & \textbf{Total} & \textbf{Base Train Set} & \textbf{Verifier Train Set} & \textbf{Test Set} \\
\hline
NAME\_STUDENT & 4394 & 1091  & 693 & 2610 \\
URL\_PERSONAL & 354  & 76    & 66  & 212 \\
EMAIL         & 112  & 29    & 21  & 62 \\
PHONE\_NUM    & 15   & 3     & 3   & 9 \\
\hline
\textbf{TOTAL} & \textbf{4871} & \textbf{1199} & \textbf{783} & \textbf{2889} \\
\hline
\end{tabular}
}
\end{table}

\subsection{Evaluation Metrics}

When evaluating the performance of a model designed to detect PII, it is essential to assess the model's ability to correctly identify true PII entities while minimizing incorrect classifications. The \textbf{True Positives} (TP) indicate the number of correctly identified PII entities. The \textbf{False Positives} (FP) represent the number of non-PII entities incorrectly classified as PII, which can reduce the utility of the dataset by unnecessarily removing valuable information. The \textbf{False Negatives} (FN) denote the number of missed detections of actual PII entities, which could result in privacy breaches and violations of legal regulations. To evaluate the model's capability in accurately detecting PII, we adopt the evaluation method suggested in \citeN{tjong2003introduction}. This method considers a PII entity to be correctly identified only if it is an exact match with the corresponding entity in the text data. To provide a comprehensive assessment of the model's performance, we consider multiple metrics that capture different aspects of the model's effectiveness.

\textbf{Precision} (Equation \ref{eq:precision}) measures the proportion of correct PII predictions out of all entities that the model classified as PII. High precision means that when the model identifies words as PII, it is very likely to be correct. This is particularly important when false positives (incorrectly labeling non-PII as PII) are costly or disruptive, such as when anonymizing educational datasets where unnecessary removal of non-PII data can reduce the value of the dataset for analysis.

\begin{equation}
\text{Precision} = \frac{TP}{TP + FP}
\label{eq:precision}
\end{equation}

\textbf{Recall} (Equation \ref{eq:recall}) measures the proportion of actual PII entities that the model successfully identifies. High recall is crucial in privacy protection, as it ensures that most, if not all, sensitive information is detected and appropriately handled, minimizing the risk of unmasked PII being exposed and resulting in potential privacy breaches or legal violations.

\begin{equation}
\text{Recall} = \frac{TP}{TP + FN}
\label{eq:recall}
\end{equation}

The \textbf{$F_{1}$ Score} (Equation \ref{eq:f1score}) provides a balance between precision and recall, offering a single metric that reflects both the model's ability to correctly identify PII and its ability to minimize false positives. The $F_{1}$ Score is especially useful when both precision and recall are equally important. 

\begin{equation}
F_{1} \text{ Score} = 2 \times \frac{\text{Precision} \times \text{Recall}}{\text{Precision} + \text{Recall}}
\label{eq:f1score}
\end{equation}

The \textbf{$F_{5}$ Score} (Equation \ref{eq:f5score}) places a stronger emphasis on recall than precision. In privacy-sensitive domains like education, where missing a piece of PII (a false negative) can be much more damaging than accidentally flagging non-PII as sensitive, the $F_{5}$ Score helps prioritize models that do a better job at catching all PII, even if it means more false positives.

\begin{equation}
F_{5} \text{ Score} = (1 + 5^2) \times \frac{\text{Precision} \times \text{Recall}}{(5^2 \times \text{Precision}) + \text{Recall}}
\label{eq:f5score}
\end{equation}

\subsection{Analysis of Cultural and Gender Bias in Name Detection by Models}

Evaluating the model's performance in detecting \texttt{<NAME\_STUDENT>} entities across different cultural and gender distributions is quite important. We adopt Hofstede's definition of culture as ``\textit{the collective programming of the mind that distinguishes the members of one group or category of people from another}'' \cite{hofstede2001cultures}. Because this collective programming shapes online expression and data creation, cultural groups are not represented equally in large-scale datasets. Training data, often sourced from the internet, tends to reflect the disproportionate influence of dominant cultures, leading to their overrepresentation. A similar imbalance exists for gender, where available data often feature a skewed distribution of male and female names. Models trained on imbalanced datasets may underperform in identifying names of groups that are underrepresented in training data. Specifically, if the training data lacks sufficient representation of names from certain cultural backgrounds, the model may exhibit a lower recall or precision for those names. This dimension of model evaluation is necessary to ensure fairness and inclusivity in PII detection systems, especially as names serve as direct identifiers with critical privacy implications.

We analyzed the cultural and gender distributions of the names in the \texttt{<NAME\_STUDENT>} entities in detail by using a two-step approach. First, we used a \textit{rule-based name parser}\footnote{\url{https://github.com/derek73/python-nameparser}} to split each name into components, typically a first name and a last name. We then determined the \textit{gender} of each name based on the first name and matched the \textit{nationality} using the last name. This method aligns with the approach described in the CRAPII paper \cite{2024.EDM-posters.88}. In the second step, we mapped the identified countries to their respective \textit{regional cultures} using the ISO-3166 dataset with the UN regional codes.\footnote{\url{https://github.com/lukes/ISO-3166-Countries-with-Regional-Codes}} Specifically, we relied on the \texttt{all.format} file, which includes detailed regional and sub-regional classifications for each country. For example, ``Nigeria'' maps to the \textit{Africa} region, while ``United States of America'' maps to the \textit{Americas}. The ISO-3166 dataset provides five cultural regions: \textit{Asia, Americas, Europe, Africa, and Oceania}. Notably, no names in the CRAPII dataset belonged to Oceania, so we focused on the remaining four cultural groups. 

To quantify potential biases, we evaluate the model for fairness using the principle of Equality of Opportunity \cite{hardt2016equalityopportunitysupervisedlearning}. This metric assesses whether the model correctly identifies student names at equal rates across different protected groups. We measure this by computing the true positive rate (TPR) for each cultural and gender subgroup, where fairness is achieved if the following condition holds: 

\[
\forall{A},P[\hat{Y}=1|A=1,Y=1] = P[\hat{Y}=1|A=0,Y=1]  
\]

Here, $Y=1$ represents a true \texttt{<NAME\_STUDENT>} entity, and $A$ is an indicator variable for membership in a protected group (e.g., $A=1$ for `Female,' $A=0$ for `Male'). This metric evaluates if the probability of a correct identification ($\hat{Y}=1$) is independent of group membership ($A$), given that a name is actually present ($Y=1$).

Our decision to aggregate countries into four continental cultural groups was driven by data sparsity. Although a country is a more granular cultural indicator, our data contains names from many countries, with most being represented by only a few samples. Therefore, we opted for this coarser grouping. This methodological choice is supported by recent research demonstrating a convergence of cultural values at the continental level \cite{jackson2024worldwidedivergence}. Similarly, our gender analysis is limited to a binary classification because, to our knowledge, there is no reliable computational method for mapping personal names to non-binary genders.

\subsection{Analysis of Models' Generalizability}
\label{generalizability}

To analyze the generalizability of our investigated models, we employed the TSCC dataset, as introduced in Section \ref{sec_data}. Notably, the TSCC dataset has been pre-redacted, with most redacted words replaced by placeholders indicating their name entity types, such as \texttt{<STUDENT>} and \texttt{<TEACHER>}. To evaluate the model's ability to generalize across diverse contexts, it was necessary to replace these placeholders with synthetic entities that reflect diversity in gender and cultural backgrounds. This replacement ensures a realistic evaluation of the models' performance when encountering unseen data with varying demographic characteristics. To generate a diverse and representative set of synthetic names, we systematically created a mapping of first and last names categorized by gender and cultural groups. This process involved the following steps:

\begin{enumerate}
    \item For each culture, we identified its corresponding countries based on the \texttt{all.format} file in the United Nations dataset.
    \item For each country, we used the \texttt{get\_top\_names} function from the \texttt{names- \\ dataset} Python package\footnote{\url{https://github.com/philipperemy/name-dataset}} to retrieve first names based on gender and last names based on the country, as outlined in the CRAPII paper \cite{2024.EDM-posters.88}.
    \item We combined the names of all the countries that belong to that culture into one list, filtering out names containing non-English characters to maintain consistency with the TSCC dataset \cite{caines-etal-2022-teacher}.
    \item Finally, the processed list of names for the given gender and cultural group was stored in a dictionary, with the group defined as a tuple of \texttt{(gender, culture)}.
\end{enumerate}

Once this mapping was complete, the 260 transcripts in the TSCC dataset were randomly assigned to the 10 gender-culture groups (2 gender by 5 culture groups), ensuring each group contained 26 transcripts. For each transcript, synthetic names were randomly sampled from the corresponding gender-culture group and used to replace placeholders. In addition to name placeholders, the dataset also contained 13 non-name placeholder categories (e.g., \texttt{〈AGE〉}, \texttt{〈DATE〉}, and \texttt{〈INSTAGRAM ACCOUNT〉}). These were replaced with synthetic entities generated using GPT-4o to ensure semantic consistency throughout the dataset.

This process resulted in a realistic and culturally diverse dataset that retains the original conversational structure while introducing diversity in entity representation. For instance, the processed version of the transcript shown in Section~\ref{sec_data} appears as follows:

\begin{verbatim}
teacher: Hi there John Doe, all OK?  
student: Hi Jane Doe, how are you?  
\end{verbatim}

\section{Results}

\begin{table}[htbp]
\centering
\caption{Performance Metrics for Different PII Detection Models (The highest values across all models are shown in bold to indicate the best performance for each metric.)}
\label{tab:all_metrics}
\renewcommand{\arraystretch}{1.5}
\resizebox{1\textwidth}{!}{%
\begin{tabular}{p{3cm}|l|rrr|rrrr}
\hline
\centering \textbf{Models} & \textbf{Entity Type} & \textbf{\# True Positive} & \textbf{\# False Positive} & \textbf{\# False Negative} & \textbf{Precision} & \textbf{Recall} & \textbf{$F_{1}$ Score} & \textbf{$F_{5}$ Score} \rule{0pt}{3.6ex}\rule[-2.2ex]{0pt}{0pt} \\ \hline

\multirow{5}{*}{\parbox{2.5cm}{\centering \textit{1. Presidio} \\ \textit{(en\_core\_web\_lg)}}} 
 & NAME\_STUDENT & 1,805 & 9,294 & 805 & 0.1626 & 0.6916 & 0.2633 & 0.6147 \\
 & URL\_PERSONAL & 181 & 2,256 & 31 & 0.0743 & 0.8538 & 0.1367 & 0.6082 \\
 & EMAIL & 61 & 10 & 1 & 0.8592 & \textbf{0.9839} & 0.9173 & 0.9784 \\
 & PHONE\_NUM & 8 & 37 & 1 & 0.1778 & \textbf{0.8889} & 0.2963 & 0.7704 \\ \cline{2-9} 
 & \textbf{Overall} & 2,055 & 11,597 & 838 & 0.1505 & 0.7103 & 0.2484 & 0.6214 \\ \hline

\multirow{5}{*}{\parbox{2.5cm}{\centering \textit{2. Presidio}\\ \textit{(en\_core\_web\_trf)}}} 
 & NAME\_STUDENT & 2,172 & 6,849 & 438 & 0.2408 & 0.8322 & 0.3735 & 0.7604 \\
 & URL\_PERSONAL & 180 & 2,257 & 32 & 0.0739 & 0.8491 & 0.1359 & 0.6049 \\
 & EMAIL & 61 & 10 & 1 & 0.8592 & \textbf{0.9839} & 0.9173 & 0.9784 \\
 & PHONE\_NUM & 8 & 37 & 1 & 0.1778 & \textbf{0.8889} & 0.2963 & 0.7704 \\ \cline{2-9} 
 & \textbf{Overall} & 2,421 & 9,153 & 472 & 0.2092 & 0.8368 & 0.3347 & 0.7503 \\ \hline

\multirow{5}{*}{\parbox{3cm}{\centering \textit{3. Azure AI Language}}} 
 & NAME\_STUDENT & 2,451 & 7,074 & 159 & 0.2573 & 0.9391 & 0.4040 & 0.8522 \\
 & URL\_PERSONAL & 145 & 917 & 67 & 0.1365 & 0.6840 & 0.2276 & 0.5926 \\
 & EMAIL & 61 & 8 & 1 & \textbf{0.8841} & \textbf{0.9839} & \textbf{0.9313} & \textbf{0.9796} \\
 & PHONE\_NUM & 8 & 161 & 1 & 0.0473 & \textbf{0.8889} & 0.0899 & 0.5279 \\ \cline{2-9} 
 & \textbf{Overall} & 2,665 & 8,160 & 228 & 0.2462 & 0.9212 & 0.3885 & 0.8333 \\ \hline

\multirow{5}{*}{\parbox{2.5cm}{\centering \textit{4. Prompting}\\ \textit{GPT-4o-mini}}} 
 & NAME\_STUDENT  & 2,036  & 750  & 574  & 0.7308  & 0.7801  & 0.7546  & 0.7781 \\
 & URL\_PERSONAL  & 153   & 313  & 59   & 0.3283  & 0.7217  & 0.4513  & 0.6899 \\
 & EMAIL          & 57    & 55   & 5    & 0.5089  & 0.9194  & 0.6552  & 0.8917 \\
 & PHONE\_NUM     & 5     & 45   & 4    & 0.1000  & 0.5556  & 0.1695  & 0.4727 \\ \cline{2-9} 
 & \textbf{Overall} & 2,251 & 1,163 & 642 & 0.6593 & 0.7781 & 0.7138 & 0.7727 \\ \hline

\multirow{5}{*}{\parbox{2.5cm}{\centering \textit{5. Fine-tuned}\\ \textit{GPT-4o-mini}}} 
 & NAME\_STUDENT  & 2,507  & 1,597  & 103  & 0.6109  & \textbf{0.9605}  & 0.7468  & \textbf{0.9398} \\
 & URL\_PERSONAL  & 199   & 206    & 13   & 0.4914  & \textbf{0.9387}  & 0.6451  & \textbf{0.9069} \\
 & EMAIL          & 60    & 10     & 2    & 0.8571  & 0.9677  & 0.9091  & 0.9630 \\
 & PHONE\_NUM     & 8     & 4      & 1    & 0.6667  & \textbf{0.8889}  & 0.7619  & 0.8776 \\ \cline{2-9} 
 & \textbf{Overall} & 2,774 & 1,817 & 119 & 0.6042  & \textbf{0.9589} & 0.7413 & \textbf{0.9377} \\ \hline

\multirow{5}{*}{\parbox{3cm}{\centering \textit{6. Verifier Model I} \\ \textit{(Without CoT)}}} 
 & NAME\_STUDENT & 2,098 & 278 & 512 & \textbf{0.8830} & 0.8038 & \textbf{0.8416} & 0.8066 \\
 & URL\_PERSONAL & 161 & 2 & 51 & \textbf{0.9877} & 0.7594 & \textbf{0.8587} & 0.7662 \\
 & EMAIL & 60 & 8 & 2 & 0.8824 & 0.9677 & 0.9231 & 0.9642 \\
 & PHONE\_NUM & 2 & 1 & 7 & 0.6667 & 0.2222 & 0.3333 & 0.2281 \\ \cline{2-9} 
 & \textbf{Overall} & 2,321 & 289 & 572 & \textbf{0.8893} & 0.8023 & \textbf{0.8435} & 0.8053 \\ \hline

\multirow{5}{*}{\parbox{3cm}{\centering \textit{7. Verifier Model II} \\ \textit{(With CoT)}}} 
 & NAME\_STUDENT & 2,261 & 704 & 349 & 0.7626 & 0.8663 & 0.8111 & 0.8618 \\
 & URL\_PERSONAL & 173 & 74 & 39 & 0.7004 & 0.8160 & 0.7538 & 0.8109 \\
 & EMAIL & 60 & 9 & 2 & 0.8696 & 0.9677 & 0.9160 & 0.9636 \\
 & PHONE\_NUM & 8 & 3 & 1 & \textbf{0.7273} & \textbf{0.8889} & \textbf{0.8000} & \textbf{0.8814} \\ \cline{2-9} 
 & \textbf{Overall} & 2,502 & 790 & 391 & 0.7600 & 0.8648 & 0.8091 & 0.8603 \\ \hline

\end{tabular}
}
\end{table}

The performance metrics of all proposed PII detection models are summarized in Table~\ref{tab:all_metrics}.

\subsection{Overall-Level Analysis of Model Performance}

\subsubsection{Presidio Models}

For the two Presidio models utilizing different \texttt{spaCy} configurations, \textit{en\_core\_web\_trf} consistently outperforms \textit{en\_core\_web\_lg} across all metrics, as shown in Table~\ref{tab:all_metrics}. The transformer-based \textit{en\_core\_web\_trf} achieves higher overall precision (0.2092 vs. 0.1505) and recall (0.8368 vs. 0.7103), likely due to its enhanced ability to capture long-range dependencies in text. However, both configurations exhibit low precision, which is likely due to the inclusive detection approach of the models. Although this approach enables the identification of a wide range of entities, including less common ones, it also leads to a significant number of false positives, thereby reducing overall precision.

\subsubsection{Azure AI Language}

The Azure AI Language model achieves an overall precision of 0.2462 and a recall of 0.9212, with the recall being the second highest across all models, slightly lower than the fine-tuned GPT-4o-mini model. Its strong recall highlights its ability to capture most true positives, reflected in the low number of false negatives (228). However, the low precision of the model, driven by a high number of false positives (8,160), limits its reliability for applications that require accurate predictions. The $F_1$ score of 0.3885 and $F_5$ score of 0.8333 further emphasize its recall-oriented nature, indicating that while Azure AI Language improves recall compared to rule-based methods, it struggles to maintain precision, resulting in an imbalanced trade-off.

\subsubsection{Prompting GPT-4o-mini}

The prompting GPT-4o-mini model achieves an overall precision of 0.6593 and recall of 0.7781, resulting in an $F_1$ score of 0.7138 and an $F_5$ score of 0.7727. Although the model demonstrates notable improvements in precision compared to rule-based approaches such as Presidio, its relatively low recall, as evidenced by the 642 false negatives, indicates that a significant number of true positives are missed. This limitation suggests that the model may not be ideal for contexts that require exhaustive PII detection. Despite these challenges, the improvement in precision highlights the potential of GPT-4o-mini's prompting capabilities, particularly for scenarios where accuracy is prioritized over comprehensive detection.

\subsubsection{Fine-tuned GPT-4o-mini}

The fine-tuned GPT-4o-mini model demonstrates strong overall performance, achieving the highest recall among all models at 0.9589. This high recall ensures that nearly all PII entities are identified, making the model highly effective for comprehensive privacy protection. Its precision of 0.6042 represents a notable improvement over both the Presidio and Azure AI Language models, highlighting the benefits of fine-tuning in balancing precision and recall. The model achieves the highest $F_5$ score (0.9377) among all models, balancing high recall with reasonable precision. This highlights the potential of fine-tuning GPT-4o-mini for PII detection in educational texts, offering clear advantages over baseline and prompting models.

\subsubsection{Verifier Models}

The \textit{Verifier Model I (Without CoT)} achieves the highest precision (0.8893) among all models by not defaulting to retaining entities when uncertain. However, this less conservative approach leads to the wrong removal of some true positives during verification, resulting in a lower recall of 0.8023. The \textit{Verifier Model II (With CoT)} demonstrates a more balanced trade-off with a recall of 0.8648, as its conservative behavior defaults to retaining entities  when uncertainty arises. Notably, the precision scores for both verifier models surpass that of all other five methods, aligning with our effort to improve precision. However, their recall is lower than that of the \textit{Fine-tuned GPT-4o-mini} model, suggesting that these verifier models may be better suited for tasks that prioritize precision over exhaustive detection.

\subsection{PII Category-Level Analysis of Model Performance}

\subsubsection{Name Detection (\texttt{NAME\_STUDENT})}

The \textit{Presidio} models demonstrate low precision (0.1626 and 0.2408) due to over-identifying common names, leading to a high number of false positives, but maintain moderate recall (0.6916 and 0.8322). The \textit{Fine-tuned GPT-4o-mini} model achieves the highest recall (0.9605) and $F_5$ score (0.9398), making it the most reliable for comprehensive name detection, despite a moderate precision of 0.6109. The \textit{Verifier Model I (Without CoT)} achieves the highest precision (0.8830) but sacrifices recall (0.8038). For names as a direct identifier, we recommend the \textit{Fine-tuned GPT-4o-mini} model, given its superior recall and $F_5$ score.

\subsubsection{URL Detection (\texttt{URL\_PERSONAL})}

For the task of detecting personal URLs (links that can reveal personal information), the \textit{Fine-tuned GPT-4o-mini} model achieves the highest recall (0.9387) and $F_5$ score (0.9069), indicating a strong performance in capturing true positives. However, its precision (0.4914) remains moderate, suggesting room for improvement in reducing false positives. The \textit{Verifier Model I (Without CoT)} significantly improves precision, achieving the highest value (0.9877) and $F_1$ score (0.8587), but at the cost of reduced recall (0.7594). This demonstrates the verifier model's effectiveness in filtering false positives while highlighting the inherent trade-off between precision and recall, as no model achieves high performance in both metrics simultaneously.

\subsubsection{Email Detection (\texttt{EMAIL})}

All models demonstrate strong performance in detecting email entities, with recall values consistently high across both rule-based and GPT-based methods. The \textit{Presidio} models and \textit{Azure AI Language} achieve the highest recall (0.9839), missing only one true email entity out of 61, followed closely by the \textit{Fine-tuned GPT-4o-mini} model and both \textit{Verifier} models with a recall of 0.9677. In terms of precision, \textit{Azure AI Language} achieves the highest value (0.8841), while the \textit{Presidio} models (0.8592), \textit{Fine-tuned GPT-4o-mini} model (0.8571), and \textit{Verifier Model I (Without CoT)} (0.8824) also perform well. In general, email detection appears to be a relatively straightforward task, with most models achieving strong performance in both recall and precision, as reflected by their high $F_1$ and $F_5$ scores.

\subsubsection{Phone Number Detection (\texttt{PHONE\_NUM})}

Five models, including both \textit{Presidio} models, \textit{Azure AI Language}, \textit{Fine-tuned GPT-4o-mini}, and \textit{Verifier Model II (With CoT)}, achieve the highest recall of 0.8889 for phone number detection. However, \textit{Presidio} and \textit{Azure AI Language} exhibit low precision, with Azure showing the lowest precision of 0.0473. In contrast, GPT-based models demonstrate higher precision, with the \textit{Fine-tuned GPT-4o-mini} model reaching 0.6667. The \textit{Verifier Model I (Without CoT)} records the lowest recall (0.2222), suggesting that it likely removed many true positives from the \textit{Fine-tuned GPT-4o-mini} model's detected entities, possibly due to the lack of Chain-of-Thought reasoning. The \textit{Verifier Model II (With CoT)} achieves the highest $F_1$ (0.8000) and $F_5$ (0.8814) scores, balancing strong precision and recall.

Overall, no single model dominates across all tested categories, as each exhibits distinct strengths. \textit{Azure AI Language} performs best for email detection, while \textit{Verifier Model II (With CoT)} is more effective for phone number detection. For names and URLs, the \textit{Fine-tuned GPT-4o-mini} model and \textit{Verifier Model I (Without CoT)} present a trade-off between recall and precision, allowing users to select a model based on the specific priorities of their task.

\subsection{Impact of Low-Precision PII Detection: Examples of Semantic Disruption}

To better understand how low-precision PII detection can disrupt the semantic integrity of datasets and hinder downstream data analysis for educational research, we present three examples. These examples demonstrate cases where \textit{Presidio} and \textit{Azure AI Language} incorrectly identify non-PII entities as PII (false positives), resulting in unnecessary replacements that alter the intended meaning of the data. Such disruptions can negatively impact the utility of the data for educational insights and analysis. In contrast, all GPT-based models successfully identify these cases as non-PII (true negatives), thereby preserving the semantic meaning and ensuring the dataset's utility for downstream research tasks.

In Example 1, \textit{Presidio} incorrectly identifies \colorbox{effort}{\textit{Jesus Christ}}, \colorbox{effort}{\textit{Mary}}, \colorbox{effort}{\textit{Joseph}}, and \colorbox{effort}{\textit{Jesus}} as PII. However, these names are not sensitive information in this context but are instead central to the story's historical and cultural narrative. The clue ``Nazareth'' is a key component of the story, as it is widely recognized as the hometown of Jesus Christ. Replacing the associated names with synthetic alternatives disrupts the educational purpose of the text, as students may no longer connect ``Nazareth'' to its religious significance. This could lead to misunderstandings and confusion in the learning process.

\vspace{2mm}
\noindent\textit{Example 1}

\noindent\textbf{Original:} \textit{At the beginning of the story, you do not know the names of the characters. Then at the end, I drop the first clue ``Nazareth'' - which is well known to be the home town of \colorbox{effort}{Jesus Christ}. You can maybe guess that the family are \colorbox{effort}{Mary} and \colorbox{effort}{Joseph} with \colorbox{effort}{Jesus} as a boy.}

\vspace{1mm}
\noindent\textbf{Replaced:} \textit{At the beginning of the story, you do not know the names of the characters. Then at the end, I drop the first clue ``Nazareth'' - which is well known to be the home town of \colorbox{outcome}{Elias Carson}. You can maybe guess that the family are \colorbox{outcome}{Lena} and \colorbox{outcome}{Daniel} with \colorbox{outcome}{Elijah} as a boy.}
\vspace{2mm}


The incorrect anonymization of names changes the intended meaning of the text, as the connection between ``Nazareth'' and its historical and religious significance is lost. This disruption affects students' understanding and the utility of the data in educational contexts. Moreover, anonymization negatively impacts the utility of the text for machine learning applications. If the anonymized text is used to train or fine-tune a language model or included in a knowledge base for a Retrieval-Augmented Generation (RAG) pipeline, it may result in incorrect or misleading responses to queries about ``Nazareth'' \cite{wang2024astuteragovercomingimperfect}. Repeated inclusion of anonymized instances, such as associating ``Nazareth'' with unrelated names like ``Elias Carson,'' may erode the model's understanding of its cultural significance, leading to inaccuracies in subsequent applications \cite{bender2021dangers,sun2019lamollanguagemodelinglifelong}.


Similarly, \textit{Azure AI Language} exhibits similar challenges when handling famous individuals. In Example 2, \textit{Azure AI Language} identifies notable figures such as \colorbox{effort}{\textit{Bill Gates}}, \colorbox{effort}{\textit{Steve Jobs}}, \colorbox{effort}{\textit{Zuckerberg}}, and \colorbox{effort}{\textit{Elon Musk}} as PII entities. The subsequent replacement with random surrogate names strips the text of its unique context and relevance. These figures' specific ages, entrepreneurial journeys, and market contexts are integral to the narrative, explaining why their stories ``\textit{didn't translate well}'' into the students' environment. 

\vspace{2mm}
\noindent\textit{Example 2}

\noindent\textbf{Original:} \textit{It became clear that while the students were excited about setting up and running startup companies on campus, they had very little background information to do so. Their role models came from the other part of the world namely \colorbox{effort}{Bill Gates}, \colorbox{effort}{Steve Jobs}, \colorbox{effort}{Zuckerberg}, \colorbox{effort}{Elon Musk} etc. Their stories or anecdotes didn’t translate well into the environment of our students.}

\vspace{1mm}
\noindent\textbf{Replaced:} \textit{It became clear that while the students were excited about setting up and running startup companies on campus, they had very little background information to do so. Their role models came from the other part of the world namely \colorbox{outcome}{Benjamin Bloom}, \colorbox{outcome}{Stephen Jackson}, \colorbox{outcome}{Oliver} \colorbox{outcome}{Underwood}, \colorbox{outcome}{Aiden Miles} etc. Their stories or anecdotes didn’t translate well into the environment of our students.}
\vspace{2mm}

Another common false positive pattern in both \textit{Presido} and \textit{Azure} is de-identifying the end and start of two consecutive sentences (with no whitespace) as a URL, illustrated by the example below:

\vspace{2mm}
\noindent\textit{Example 3}

\noindent\textbf{Original:} \textit{ Consider hiring a copywriter to craft a compelling \colorbox{effort}{menu.K}eep menus clean – no grease and no food or water stains. Get rid of worn or torn menus.Update menu and prices at least once a \colorbox{effort}{year.Build} menu around popular items. }

\vspace{1mm}
\noindent\textbf{Replaced:} \textit{Consider hiring a copywriter to craft a compelling \\ \colorbox{outcome}{https://techwaveinsight.io}eep menus clean – no grease and no food or water stains. Get rid of worn or torn menus.Update menu and prices at least once a \colorbox{outcome}{http://elitecodingacademy.org} menu around popular items.}
\vspace{2mm}

Replacing key terms such as ``\textit{menu},'' ``\textit{Keep},'' ``\textit{year},'' and ``\textit{Build}'' with arbitrary URLs disrupts the original meaning. The instructions lose clarity, merging steps into a single ill-structured, confusing sentence. 

\subsection{Cost Analysis}

\begin{table}[ht]
\centering
\caption{Cost Breakdown for PII Detection Models (in USD)}
\vspace{0.4cm}
\label{tab:cost}
\renewcommand{\arraystretch}{1.7}
\resizebox{0.9\textwidth}{!}{%
\large
\begin{tabular}{|p{6.9cm}|>{\centering\arraybackslash}p{2cm}|>{\centering\arraybackslash}p{2.5cm}|>{\centering\arraybackslash}p{3cm}|>{\centering\arraybackslash}p{2.5cm}|>{\centering\arraybackslash}p{2.5cm}|>{\centering\arraybackslash}p{2.5cm}|>{\centering\arraybackslash}p{2.5cm}|}
\hline
\centering \textbf{Models} & \textbf{Base Fine-tuning} & \textbf{Base Model Dependency} & \textbf{Verifier Training Data Construction} & \textbf{Verifier Fine-tuning} & \textbf{Evaluation} & \textbf{Total} & \textbf{Average (per 1M tokens)} \\
\hline
\textit{\large Presidio (en\_core\_web\_lg)} & \multicolumn{4}{c|}{\large \textemdash} & \large 0 & \large 0 & \large 0 \\
\textit{\large Presidio (en\_core\_web\_trf)} & \multicolumn{4}{c|}{\large \textemdash} & \large 0 & \large 0 & \large 0 \\
\textit{\large Azure AI Language} & \multicolumn{4}{c|}{\large \textemdash} & \large \$63.27 & \large \$63.27 & \large \$4.90 \\
\textit{\large Prompting GPT-4o-mini} & \multicolumn{4}{c|}{\large \textemdash} & \large \$5.22 & \large \$5.22 & \large \$0.40 \\ \hline
\textit{\large Fine-tuned GPT-4o-mini} & \large \$7.22 & \multicolumn{3}{c|}{\large \textemdash} & \large \$4.71 & \large \$11.93 & \large \$0.92 \\ \hline
\textit{\large Verifier Model I (Without CoT)} & \large \textemdash & \large \$11.93 & \large \$1.16 & \large \$0.80 & \large \$0.08 & \large \$13.97 & \large \$1.09 \\
\textit{\large Verifier Model II (With CoT)} & \large \textemdash & \large \$11.93 & \large \$1.50 & \large \$1.08 & \large \$0.18 & \large \$14.69 & \large \$1.13 \\
\hline
\end{tabular}
}
\end{table}

Table~\ref{tab:cost} presents the cost associated with each PII detection model, complementing the performance metrics in Table~\ref{tab:all_metrics}. The results emphasize the potential of GPT-based approaches, particularly the \textit{Fine-tuned GPT-4o-mini} model, for high-quality PII detection at significantly lower costs.

The \textit{Fine-tuned GPT-4o-mini} model achieves the highest overall recall (0.9589) and $F_5$ score (0.9377), outperforming both the free \textit{Presidio} models and the expensive \textit{Azure AI Language} model. While the \textit{Presidio} models incur no cost, their low precision (0.1505 and 0.2092) and $F_5$ scores (0.6214 and 0.7503) highlight their limitations in balancing false positives and true positives. In contrast, the \textit{Azure AI Language} model, though more precise, costs \$63.27 (\$4.90 per 1M tokens), which is approximately 6 times higher than the \textit{Fine-tuned GPT-4o-mini} model (\$0.92 per 1M tokens). Then, the \textit{Verifier models} marginally increase the total cost to \$13.97 (Without CoT) and \$14.69 (With CoT), enhancing precision for applications where minimizing false positives is critical. Despite the additional cost, these models remain far more economical than \textit{Azure AI Language} model while retaining GPT's high recall and semantic accuracy. Thus, GPT-based models, led by the \textit{Fine-tuned GPT-4o-mini} model, outperform both \textit{Presidio} and \textit{Azure AI Language} in balancing cost and performance. This underscores their potential as an efficient and scalable solution for PII detection in educational data.

\subsection{Name Culture and Gender Bias Analysis}
Based on the results in Table~\ref{tab:all_metrics}, we selected three models\textemdash \textit{Presidio with en\_core\_web\_trf}, \textit{Azure AI Language}, and \textit{Fine-tuned GPT-4o-mini}\textemdash for the analysis of cultural and gender bias in name detection. To statistically validate these observations, we also perform two-tailed Mann-Whitney U tests. Each test evaluates whether a model's performance on a specific protected group is significantly different from its performance on the aggregate of the other groups within the same category (e.g., recall on African names vs. all non-African names). The total number of entities for each gender and culture group and their recall are shown in Table~\ref{tab:recall_comparison}. The results of these tests are shown in Table~\ref{tab:statistical_significance}.

\begin{table}[htbp]
\centering
\renewcommand{\arraystretch}{1.3}
\caption{Recall comparison across gender and culture groups for selected models}
\vspace{0.4cm}
\label{tab:recall_comparison}
\resizebox{0.6\textwidth}{!}{
\begin{tabular}{|l|l|r|r|r|r|}
\hline
\textbf{Type} & \textbf{Group} & \textbf{Total} & \textbf{Presidio} & \textbf{Azure AI} & \textbf{GPT-4o-mini} \\ \hline
\multirow{2}{*}{Gender} & Male       & 1582          & 0.8786         & 0.9494          & 0.9646          \\ \cline{2-6} 
                        & Female     & 1002          & 0.8832         & 0.9541          & 0.9591          \\ \hline
\multirow{4}{*}{Culture} & Europe        & 410           & 0.9024         & 0.9585          & 0.9756          \\ \cline{2-6} 
                        & Americas       & 858           & 0.9091         & 0.9755          & 0.9790          \\ \cline{2-6} 
                        & Asia           & 500           & 0.8640         & 0.9320          & 0.9840          \\ \cline{2-6} 
                        & Africa         & 238           & 0.7647         & 0.9244          & 0.9748          \\ \hline
\end{tabular}
}
\end{table}

\begin{table}[htbp]
\centering
\small
\renewcommand{\arraystretch}{1.3}
\caption{Statistics on TPR differences across gender and cultural groups for selected models, as measured by the Mann-Whitney U test}
\vspace{0.3cm}
\label{tab:statistical_significance}
\resizebox{0.71\textwidth}{!}{%
\begin{tabular}{|l|l|l|r|r|r|l|}
\hline
\textbf{Category} & \textbf{Group} & \textbf{Model} & \textbf{Group TPR} & \textbf{Rest TPR} & \textbf{U-statistic} & \textbf{p-value} \\ \hline
\multirow{6}{*}{Gender} & \multirow{3}{*}{Male} 
    & Presidio     & 0.8786 & 0.8832 & 788937 & n.s. \\ \cline{3-7}
 &  & Azure        & 0.9494 & 0.9541 & 788888 & n.s. \\ \cline{3-7}
 &  & GPT-4o-mini  & 0.9646 & 0.9591 & 796957 & n.s. \\ \cline{2-7}
 & \multirow{3}{*}{Female} 
    & Presidio     & 0.8832 & 0.8786 & 796431 & n.s. \\ \cline{3-7}
 &  & Azure        & 0.9541 & 0.9494 & 796480 & n.s. \\ \cline{3-7}
 &  & GPT-4o-mini  & 0.9591 & 0.9646 & 788411 & n.s. \\ \hline
\multirow{12}{*}{Culture} & \multirow{3}{*}{Europe} 
    & Presidio     & 0.9024 & 0.8734 & 336670 & n.s. \\ \cline{3-7}
 &  & Azure        & 0.9585 & 0.9543 & 328579 & n.s. \\ \cline{3-7}
 &  & GPT-4o-mini  & 0.9756 & 0.9799 & 325760 & n.s. \\ \cline{2-7}
 & \multirow{3}{*}{Americas} 
    & Presidio     & 0.9091 & 0.8571 & 518076 & \textbf{***} \\ \cline{3-7}
 &  & Azure        & 0.9755 & 0.9399 & 510039 & \textbf{***} \\ \cline{3-7}
 &  & GPT-4o-mini  & 0.9790 & 0.9791 & 492456 & n.s. \\ \cline{2-7}
 & \multirow{3}{*}{Asia} 
    & Presidio     & 0.8640 & 0.8845 & 368796 & n.s. \\ \cline{3-7}
 &  & Azure        & 0.9320 & 0.9628 & 364898 & \textbf{***} \\ \cline{3-7}
 &  & GPT-4o-mini  & 0.9840 & 0.9774 & 378976 & n.s. \\ \cline{2-7}
 & \multirow{3}{*}{Africa} 
    & Presidio     & 0.7647 & 0.8948 & 183022 & \textbf{***} \\ \cline{3-7}
 &  & Azure        & 0.9244 & 0.9593 & 203048 & \textbf{**} \\ \cline{3-7}
 &  & GPT-4o-mini  & 0.9748 & 0.9796 & 209372 & n.s. \\ \hline
\multicolumn{7}{l}{\small \textit{Significance levels:} n.s. = not significant, \textbf{*} $p<0.1$, \textbf{**} $p<0.05$, \textbf{***} $p<0.01$} \\
\end{tabular}%
}
\end{table}

\textbf{Gender Analysis.} The results indicate that the recall scores between male and female names across the three models are similar. For \textit{Presidio} and \textit{Azure AI Language}, there is a marginally higher recall for female names compared to male names. Specifically, \textit{Azure AI Language} achieves a recall of 0.9541 for female names and 0.9494 for male names. In contrast, the \textit{Fine-tuned GPT-4o-mini} model performs slightly better on male names (0.9646) compared to female names (0.9591). However, the differences in recall for male and female names are minimal, suggesting consistent performance between gender groups in all models.

\textbf{Culture Analysis.} The cultural analysis reveals more significant differences in model performance. Both Microsoft models show significantly lower recall for African and Asian names. In particular, \textit{Presidio} exhibits a notable performance gap, with recall rates of 0.7647 for African names and 0.8640 for Asian names, compared to 0.9024 for European and 0.9091 for American names. \textit{Azure AI Language} also shows a lower recall for African names (0.9244), although the gap is less pronounced than in \textit{Presidio}. These results suggest inherent biases in the Microsoft models, possibly stemming from imbalances in the training data or gaps in cultural representation. In contrast, the \textit{Fine-tuned GPT-4o-mini} model achieves consistent and high recall across all cultural groups, demonstrating no statistically significant performance differences among the cultural groups. It performs with recall scores of 0.9756 for European names, 0.9790 for American names, 0.9840 for Asian names, and 0.9748 for African names. This minimal variation in recall across cultural groups demonstrates the model’s ability to mitigate cultural bias and generalize effectively across diverse name distributions.

Therefore, the gender analysis does not show significant differences in recall for male and female names across all models, indicating consistent performance in this aspect. However, cultural analysis reveals that both \textit{Presidio} and \textit{Azure AI Language} exhibit performance gaps for African and Asian names, with lower recall for these groups. The \textit{Fine-tuned GPT-4o-mini} model, on the other hand, performs generally well across all cultural groups, effectively addressing the bias observed in the baseline models and highlighting its superior generalization capability.

\subsection{Generalizability Analysis}

\begin{table}[ht]
\centering
\caption{Performance Metrics for Different PII Detection Approaches on the TSCC Dataset (The highest values across all models are shown in bold to indicate the best performance for each metric.)}
\vspace{0.4cm}
\label{tab:tscc_metrics}
\renewcommand{\arraystretch}{1.5}
\resizebox{0.9\textwidth}{!}{%
\large
\begin{tabular}{p{11cm}|rrr|rrrr}
\hline
\centering \textbf{Models} & \textbf{\# TP} & \textbf{\# FP} & \textbf{\# FN} & \textbf{Precision} & \textbf{Recall} & \textbf{$F_{1}$ Score} & \textbf{$F_{5}$ Score} \rule{0pt}{3.6ex}\rule[-2.2ex]{0pt}{0pt} \\ \hline

\textit{1. Presidio (en\_core\_web\_trf)} 
& 1,513 & 2,694 & 102 & 0.3596 & 0.9368 & 0.5198 & 0.8824 \\ \hline

\textit{2. Azure AI Language} 
& 1,320 & 1,316 & 295 & 0.5008 & 0.8173 & 0.6210 & 0.7979 \\ \hline

\textit{3. GPT-4o-mini + few-shot prompting} 
& 1,604 & 641 & 11 & 0.7145 & \textbf{0.9932} & 0.8311 & 0.9785 \\ \hline

\textit{4. Fine-tuned GPT-4o-mini + zero-shot prompting} 
& 1,273 & 2 & 342 & \textbf{0.9984} & 0.7882 & 0.8810 & 0.7947 \\ \hline

\textit{5. GPT-4o-mini + fine-tuning} 
& 1,561 & 26 & 54 & 0.9836 & 0.9666 & 0.9750 & 0.9672 \\ \hline

\textit{6. Fine-tuned GPT-4o-mini + fine-tuning} 
& 1,598 & 48 & 17 & 0.9708 & 0.9895 & \textbf{0.9801} & \textbf{0.9887} \\ \hline

\end{tabular}
}
\end{table}

To further evaluate the performance of different models for PII detection, we further used the TSCC dataset (as introduced in Section \ref{sec_data}) to evaluate the models that were implemented on the CRAPII dataset. Table~\ref{tab:tscc_metrics} presents the performance for different PII detection models on the TSCC dataset. The metrics are the same as those presented in Table~\ref{tab:all_metrics}.

We selected one Presidio model \textit{Presidio (en\_core\_web\_trf)} model as it demonstrated higher precision and recall compared to the \textit{en\_core\_web\_lg} variant (Model 1 from Table~\ref{tab:all_metrics}). As shown in Table \ref{tab:tscc_metrics}, the \textit{Presidio (en\_core\_web\_trf) model} demonstrates relatively high recall (0.9368) and low precision (0.3596). This is likely due to its conservative approach to entity detection, which enables it to capture a wide range of entities, including rare or unconventional names. This conservatism results in a large number of false positives (2,694), the highest among all six models, which significantly impacts its precision.

Then, \textit{Azure AI Language} was chosen as another baseline to compare with both rule-based methods and LLM-based models. The  \textit{Azure AI Language} model exhibits relatively balanced precision (0.5008) and recall (0.8173). However, its overall performance is suboptimal compared to the GPT-based approaches. With 1,316 false positives and 295 false negatives, the model fails to achieve the level of precision or recall seen in other approaches. This is reflected in its moderate $F_{1}$ score of 0.6210 and $F_{5}$ score of 0.7979.

Next, we investigated the \textit{Fine-tuned GPT-4o-mini} model which was fine-tuned on the CRAPII dataset. To gain a better understanding of the GPT models on PII detection on TCSS dataset. We used a three-shot prompting strategy to directly prompted GPT-4o-mini to identify PII entities in the TSCC dataset.  We also prompted the \textit{Fine-tuned GPT-4o-mini} model (fine-tuned on the CRAPII dataset, Model 5 in Table~\ref{tab:all_metrics}) without examples to assess its generalizability on the unseen TSCC dataset.  Both models employed the same prompt structure presented in Table~\ref{table:prompt_structures}, adjusting the user prompt to only label names with special identifiers as names were the only PII category in the dataset. The results in Table \ref{tab:tscc_metrics} show that directly prompting GPT-4o-mini with few-shot examples achieves the highest recall among all models at 0.9932, demonstrating the model’s capability to detect almost all true PII entities in the dataset. However, its precision is relatively low at 0.7145 due to the higher number of false positives (641). While this approach achieves a strong $F_{5}$ score of 0.9785, the $F_{1}$ score of 0.8311 indicates that the lower precision affects its overall performance. Then, prompting the \textit{Fine-tuned GPT-4o-mini} model (on the CRAPII dataset) results in the highest precision across all models (0.9984), with only two false positives. This indicates that the entities it detects are highly accurate. However, the recall is relatively low at 0.7882, leading to a $F_{1}$ score of 0.8810 and a lower $F_{5}$ score of 0.7947. This approach is particularly suited for scenarios where precision is more critical than recall.

We also provided users with a practical option to improve model performance by fine-tuning on a minimal labeled subset of their dataset, enabling the model to better align with the specific characteristics of the TSCC dataset. Of the 260 transcripts in the processed TSCC dataset, we randomly selected 10 transcripts for fine-tuning purposes introduced below and used the remaining 250 transcripts for evaluation across all models. First, we directly fine-tuned GPT-4o-mini on the 10 selected transcripts. Second, we further fine-tuned the \textit{Fine-tuned GPT-4o-mini} model (Model 5 from Table~\ref{tab:all_metrics}) using the same 10 transcripts. Both fine-tuning models used the prompt structure described in Table~\ref{table:fine_tuning_prompt_structures}. As with prompting, the only adjustment was to instruct the model to label names using \texttt{@@@} and \texttt{\#\#\#}. Fine-tuning GPT-4o-mini on the 10 selected TSCC transcripts achieves a strong balance between precision (0.9836) and recall (0.9666). With only 26 false positives and 54 false negatives, this model achieves a $F_{1}$ score of 0.9750 and a $F_{5}$ score of 0.9672. This demonstrates the potential of fine-tuning even on a small labeled subset to adapt the model to new datasets effectively. Then, 
further fine-tuning of the \textit{Fine-tuned GPT-4o-mini} model (on the CRAPII dataset) using the 10 selected transcripts results in the best overall performance. This model achieves a high recall of 0.9895 and a precision of 0.9708, striking an excellent balance between the two metrics. Although its recall is slightly lower (about 0.3\%) than the few-shot prompting model, its precision improves significantly by approximately 26\%. This leads to the highest $F_{1}$ score (0.9801) and the $F_{5}$ score (0.9887) among all models, making it the most robust approach to PII detection on the TSCC dataset.

Therefore, the \textit{Fine-tuned GPT-4o-mini + fine-tuning} approach achieves the highest $F_{1}$ and $F_{5}$ scores, highlighting its ability to maintain high precision and recall simultaneously. These results also demonstrate the potential of GPT-based approaches over traditional models such as \textit{Presidio} and \textit{Azure}, offering superior performance in both precision and recall. While other approaches, such as few-shot prompting with GPT-4o-mini, excel in recall, the overall balance achieved by fine-tuning makes it a versatile option. Users can choose the most appropriate method based on their specific requirements, whether to prioritize recall, precision, or a combination of both.

\section{Discussion and Conclusion}


Our study demonstrates notable performance and cost-effectiveness of a \textit{Fine-tuned GPT-4o-mini} model for identifying personally identifiable information in educational texts. Compared to established rule-based (\textit{Presidio}) and commercial (\textit{Azure AI Language}) methods, our fine-tuned model achieves a recall of over 95\%, providing a viable approach for privacy protection. To address the need for high precision in downstream applications (e.g., curating anonymized datasets to support machine learning and educational research), our study also introduced a verifier model capable of filtering false positives, thus helping to preserve the semantic meaning of data when using methods such as Hidden in Plain Sight.

Furthermore, our fairness analysis reveals that the \textit{Fine-tuned GPT-4o-mini} model delivers balanced performance across cultural and gender groups, reducing the biases present in baseline systems. Testing on a unseen, different data corpus confirms the model's ability to generalize, as it requires only limited fine-tuning to achieve similar performance. Finally, our cost analysis shows that this strong performance is achievable at a lower cost than other commercial services, reducing barriers to privacy-preserving research in education.

\subsection{Considerations for Deploying Proprietary and Open-Source Models}

\textbf{Risk of PII leakage.} While our study demonstrates notable performance of using the proprietary model, i.e., GPT-4o-mini from OpenAI, the handling of sensitive data raises relevant considerations. Since data must be transmitted to OpenAI's servers, there is a possible risk that PII, if used in training, could be exposed through techniques like prompt injection attacks, as highlighted by recent studies \cite{nasr2023scalableextractiontrainingdata}. However, it should be noted that OpenAI has implemented data processing agreements and maintains a policy ensuring that data submitted via their API is not used for model training.\footnote{\url{https://openai.com/policies/data-processing-addendum}} These measures provide an added layer of safeguard, reducing the risk of unintentional data leakage. In comparison, while deploying open-source models locally ensures that no third-party data processors gain access to PII, it still carries baseline privacy risks. Open-source models downloaded from platforms like Hugging Face may contain obfuscated harmful code that executes during model use, allowing attackers to perform reverse shell attacks to gain control of the user’s machine and access PII, leading to further damage \cite{Zhao_2024}.


\textbf{Implementation Cost.} Open-source LLMs (e.g.,  LLaMA 3, Mistral, DeepSeek, etc.) give developers full control over their data and fine-tuning process, making them a suitable option for privacy-sensitive tasks such as educational PII de-identification. However, adopting these models also introduces implementation challenges. A typical 8-billion-parameter model (e.g., LLaMA 3.1 8B \cite{llama3_2024}) needs more than 90 GB of GPU memory to fine-tune—beyond what a single GPU can provide. This constraint pushes teams toward multi-GPU clusters and distributed training, both of which require specialized skills in parallel computing. Although memory saving methods such as LoRA and ZeRO can reduce the requirement to roughly 6 GB of VRAM, configuring them correctly still requires technical knowledge in large-model optimization \cite{singh2024studyoptimizationsfinetuninglarge}. In short, open-source models are scalable and secure once deployed, but they are best suited for teams with sufficient hardware resources and technical experience. In comparison, commercial APIs such as GPT and Claude offer easier deployment and minimal infrastructure requirements, making them more accessible for small teams or educational institutions.

\textbf{Energy footprint.}
Beyond implementation costs, the operational energy consumption of LLMs raises  environmental concerns due to the associated carbon footprint. Measuring the energy footprint is challenging for proprietary models due to limited public information on model architecture and deployment, making it difficult to assess the energy costs borne by providers. While direct measurement of energy consumption is difficult, many AI models are deployed on cloud computing platforms, which allows for indirect estimation. A study measuring the energy consumption of various models found that GPT-4o-mini is among the most efficient of OpenAI's models, with an energy cost comparable to open-source LLaMA models \cite{jegham2025hungryaibenchmarkingenergy}. This study noted that lower electricity usage does not always equal to superior energy efficiency: models that cost less energy to run may perform worse. This distinction matters in educational PII detection, as prior work confirms that LLaMA 3.1 8B achieves suboptimal performance, particularly in terms of precision \cite{ctx11692351300004436}. Therefore, a model like GPT-4o-mini may be a reasonable choice, balancing a manageable energy footprint with the high accuracy required for effective PII de-identification.

\subsection{Implications}


\subsubsection{Integrating Privacy-Preserving Models in AI Educational Platforms}

Our study explores potential solutions to the rising concerns about privacy and data security in the deployment of AI-powered educational technologies \cite{ive2024privacypreservingbehaviourchatbotusers}, such as chatbots. We recommend that future developments of educational tools—particularly those integrating chatbots into teaching and learning environments—consider adopting privacy-preserving models within their system development processes. Many teachers and students may unknowingly expose sensitive personal information—such as their name, age, gender, and occupation—when interacting with these AI-driven platforms \cite{ive2024privacypreservingbehaviourchatbotusers}. This is especially likely when users are prompted to share such information in exchange for more personalized responses from the chatbot \cite{carmichael2022users}, a common strategy in prompt engineering to enhance the accuracy of AI outputs.
Moreover, as some educational tools (e.g., chatbots) collect data for performance enhancement or research purposes, there is a risk that this data could be further exposed or misused \cite{nasr2023scalableextractiontrainingdata}. Given the increasing reliance on data for improving AI systems, there is a growing need to support safer data handling practices in educational tools. In such cases, privacy-preserving techniques—such as using a PII detection model like the one explored in this study in combination with the Hidden in Plain Sight method—can help replace detected PII with synthetic entities that preserve the semantic structure of the original content. This approach may enable broader use of AI tools, including chatbots, in educational contexts without exposing sensitive personal information.


\subsubsection{Expanding the Evaluation of PII De-Identification in Educational Data Mining}

Our findings suggest that relying solely on aggregate performance metrics may be insufficient for thoroughly evaluating PII de-identification systems. Such metrics can obscure underlying flaws in a model’s behavior. We observed that these limitations can have ethical and practical implications, particularly in educational settings where fairness is important. To address this, we recommend that researchers and practitioners consider incorporating additional evaluation strategies—such as (1) performance breakdowns by demographic subgroups to assess fairness (e.g., equality of odds and equal opportunity), and (2) evaluations on unseen datasets to evaluate generalizability. By adopting more comprehensive evaluation practices, the reliability and fairness of PII de-identification systems can be improved for real-world educational applications.

\subsection{Future Works}

\textit{First}, our evaluation primarily addressed PII categories that are commonly recognized by existing baseline methods (e.g., names, emails). Extending this approach to more granular or domain-specific categories—such as street addresses, student ID numbers, or indirect identifiers like dates and ages—poses additional challenges. Identifying these categories often requires deeper contextual understanding; for example, a string of digits could represent an ID number or an unrelated quantity, and an age is only sensitive when associated with an identifiable individual. Successfully detecting such PII would require the creation of new, high-quality annotated datasets and the development of more context-aware fine-tuning strategies.

\textit{Second}, while we focused on Microsoft Presidio and Azure AI Language for benchmarking due to their prevalence, a broader comparative analysis would be valuable for future work. Incorporating additional frameworks—such as Google’s Cloud Data Loss Prevention API, Amazon Comprehend, or open-source models from platforms like Hugging Face—could help further clarify the strengths and limitations of our approach. Such comparisons may also highlight performance-cost trade-offs and reveal cases where alternative models perform better for specific data types.

\textit{Third}, although we conducted a preliminary generalizability test on the TSCC dataset, the diverse nature of educational contexts suggests that some unseen PII distributions may still pose challenges. A more comprehensive evaluation across a broader range of datasets, representing different age groups (K-12 vs. higher education) and context (e.g., lecture or student presentation), is needed to better define the model’s operational boundaries and identify where domain-specific fine-tuning may be necessary.

\textit{Fourth}, while the verifier models we developed enhance precision, their trade-off in recall requires further improvement. The current implementation, although effective at reducing false positives, can overcorrect by eliminating true PII entities, which poses a risk in privacy-sensitive applications. Future work could explore alternative multistage verification strategies, such as implementing the verification process as a multi-model plurality or veto vote, to further reduce PII leakage. This may enable more nuanced filtering, preserving high recall while improving precision.

\textit{Fifth}, although this study focused on optimizing detection and verification processes, incorporating red-teaming or adversarial testing strategies is a promising direction for future work. These methods can help uncover blind spots and edge cases that may be overlooked during standard evaluation, particularly in noisy, multilingual, or informal educational data. By simulating targeted attack scenarios or modeling real-world misuse patterns, adversarial testing can offer valuable stress tests of model assumptions and reveal subtle failure modes. Integrating such strategies into the development pipeline could improve the robustness and reliability of PII detection systems across diverse deployment contexts.

\textit{Sixth}, despite their high precision, fine-tuned GPT-4o-mini models exhibit false alarms for a specific group of non-PII entities: fictional names in student essays. For example, the name \colorbox{effort}{``Rick"} in ``Personas are based on user research but tell a story about your insights. An example persona might be \colorbox{effort}{Rick}, a 47-year-old manager struggling with his work-family-life balance" is a fictional persona from a student's essay, yet it is incorrectly identified as PII. These errors suggest that certain fictional or example names can confuse high-precision PII detection systems. Future work could focus on better distinguishing these cases, which are especially common in educational texts. Additionally, fine-tuned GPT-4o-mini sometimes fails to detect actual PII, such as author names. For example, \colorbox{effort}{``Maribel Navarrete"} in ``Francesco 1 \colorbox{effort}{Maribel Navarrete} Design Thinking for Business Innovation April 21" is not identified as PII. This highlights the need for more context-aware training strategies to improve the model’s handling of ambiguous or domain-specific PII.


\textit{Finally}, beyond the precision–recall trade-off, our model’s robustness is limited by domain shift. When evaluated on the out-of-domain TSCC dataset, which consists of teacher–student conversations, the model showed a notable drop in recall. This performance degradation is likely due to distributional differences between TSCC and our training corpus, CRAPII, which comprises student essays. To build a more broadly applicable PII detection model, future work can prioritize fine-tuning on a larger, more diverse dataset that reflects a wider range of educational contexts. This could improve the model’s generalizability and reduce performance loss when applied to unseen domains.


\section*{Declaration of Generative AI Software Tools in the Writing Process}
During the preparation of this work, the authors used ChatGPT-4o in all sections in order to polish up the language and correct any grammatical errors. After using this tool, the authors reviewed and edited the content as needed and take full responsibility for the content of the publication.

\section*{Acknowledgments}
This research was funded by the the Learning Engineering Virtual Institute (\href{https://learning-engineering-virtual-institute.org/}{https://learning-engineering-virtual-institute.org/}) and Richard King Mellon Foundation (Grant \#10851). The opinions, findings, and conclusions expressed in this paper are those of the authors alone. The authors extend their sincere gratitude to Yuting Wang, Andra Liu, Owen Lalis, and Shivang Gupta for their  invaluable insights and suggestions on the initial development of our study.

\bibliographystyle{acmtrans}
\bibliography{ref}

\end{document}